%% file: egpaper_for_review.tex
\documentclass[10pt,twocolumn,letterpaper]{article}

\usepackage{wacv}
\usepackage{times}
\usepackage{epsfig}
\usepackage{graphicx}
\usepackage{amsmath}
\usepackage{amssymb}
\usepackage{multirow}
\usepackage{algpseudocode}
\usepackage{algorithm}
\usepackage[table]{xcolor}
\usepackage{subfig}

\wacvfinalcopy 

\def\wacvPaperID{250} 

\ifwacvfinal\fi
\begin{document}

\title{Unsupervised Learning of Camera Pose with Compositional Re-estimation}


\author{Seyed Shahabeddin Nabavi\\
York University\\
{\tt\small nabaviss@yorku.ca}
\and
Mehrdad Hosseinzadeh\\
University of Manitoba\\
{\tt\small mehrdad@cs.umanitoba.ca }
\and
Ramin Fahimi\\
Ryerson University\\
{\tt\small ramin.fahimi@ryerson.ca}
\and
Yang Wang\\
University of Manitoba\\
{\tt\small ywang@cs.umanitoba.ca}
}

\maketitle
\ifwacvfinal\thispagestyle{empty}\fi

\input{Abstract_V2}
\input{Introduction_V6}
\input{RelatedWork_V2}
\input{Method_V6_SH}
\input{Experiment_V2}
\input{Conclusion}
\input{aknowledegment}
\newpage

{\small
\bibliographystyle{ieee}
\bibliography{egbib}
}

\end{document}

%% file: Abstract_V2.tex
\begin{abstract}
We consider the problem of unsupervised camera pose estimation. Given an input video sequence, our goal is to estimate the camera pose (i.e. the camera motion) between consecutive frames. Traditionally, this problem is tackled by placing strict constraints on the transformation vector or by incorporating optical flow through a complex pipeline. We propose an alternative approach that utilizes a compositional re-estimation process for camera pose estimation. Given an input, we first estimate a depth map. Our method then iteratively estimates the camera motion based on the estimated depth map. Our approach significantly improves the predicted camera motion both quantitatively and visually. Furthermore, the re-estimation resolves the problem of out-of-boundaries pixels in a novel and simple way. Another advantage of our approach is that it is adaptable to other camera pose estimation approaches. Experimental analysis on KITTI benchmark dataset demonstrates that our method outperforms existing state-of-the-art approaches in unsupervised camera ego-motion estimation.
\end{abstract}

%% file: Introduction_V6.tex
\section{Introduction}

We tackle the problem of visual odometry (VO), where the goal is to estimate the camera poses (e.g. motion) given a number of consecutive frames in a video sequence. This problem plays an important role in many real-world applications, such as self-driving vehicles \cite{chen2015deepdriving}, obstacle avoidance \cite{nister2006visual}, interactive robots \cite{fong2003survey} and navigation systems \cite{fraundorfer2007topological}. In the presence of a single RGB camera (i.e. monocular), this problem has been explored in \cite{zhou2017unsupervised,Yin_2018_CVPR,mahjourian2018unsupervised,zhan2018unsupervised,luo2018every,costante2018ls,li2018undeepvo,li2018undeepvo,konda2015learning,wang2017deepvo} from various perspectives and under different assumptions. Our work is particularly inspired by a recent line of work~\cite{zhou2017unsupervised,Yin_2018_CVPR,mahjourian2018unsupervised} on learning monocular camera pose estimation and depth estimation in an \textit{unsupervised} setting. The only available data in this setting during training are monocular frames and camera intrinsics. The model is learned to map the input pixels to an estimate of camera poses (parameterized as transformation matrices) and scene structures (parameterized as depth maps). During testing, the input to the model is the raw video. We will use the learned model to produce the camera poses of the test video. As a by-product, we will also obtain the predicted depth map on each frame of the test video.

\begin{figure}
  \centering
  \includegraphics[width=0.97\linewidth]{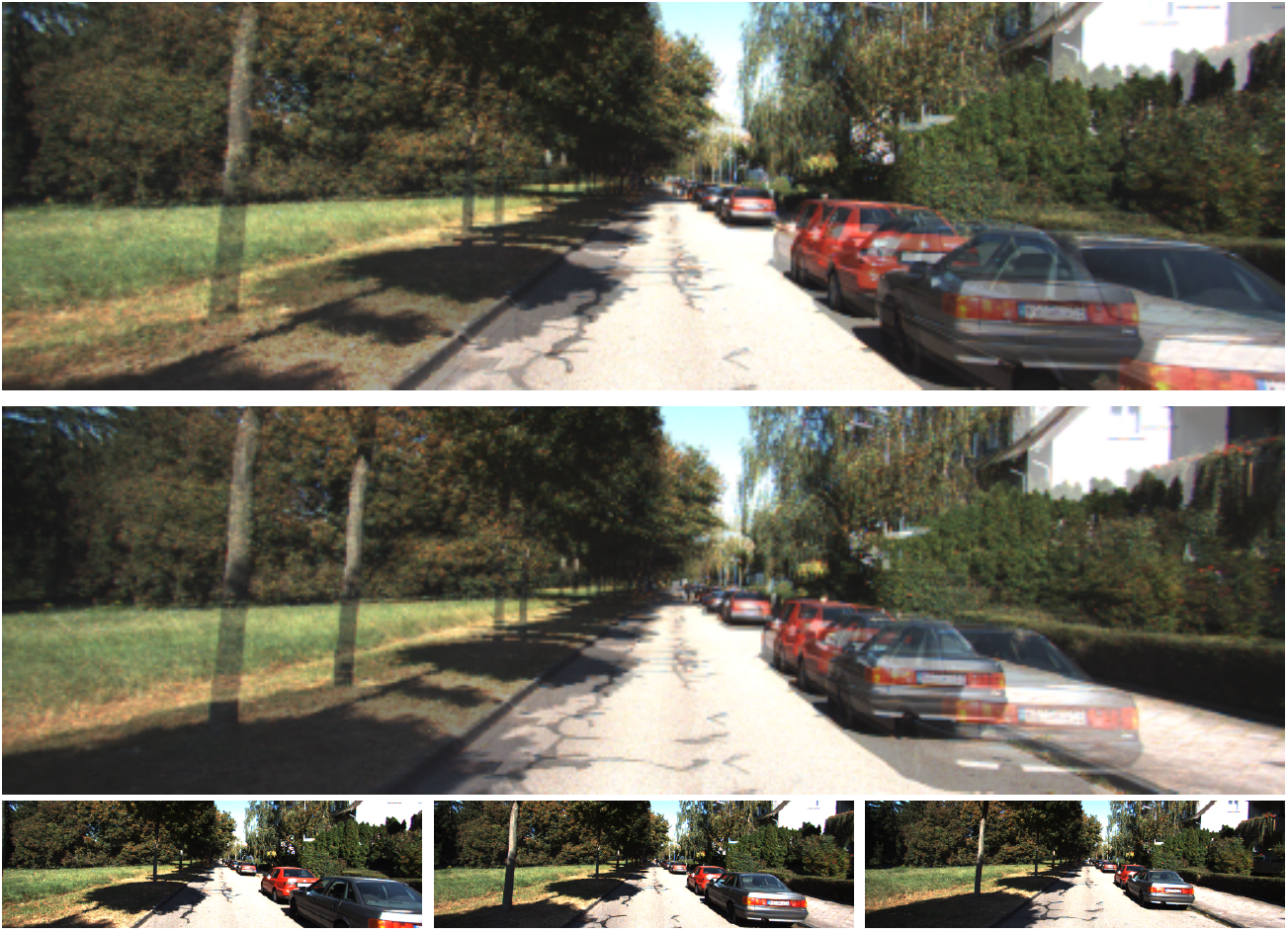}
  \caption{An illustration of the problem of large displacement between two views in pose estimation with the view synthesis formulation. The 3rd row shows three consecutive frames in a video. The 1st row shows the difference between the left and middle frames. The 2nd row shows the difference between the middle and right frames. When the displacement of two views is large, the assumption made by the view synthesis no longer holds. In this paper, we propose an alternative approach that splits the estimation into smaller pieces and re-estimate the transformation through a compositional transformation estimation.}
  \label{fig:intro}
\end{figure}

Several previous works (e.g. \cite{zhou2017unsupervised,Yin_2018_CVPR,mahjourian2018unsupervised}) have been proposed to estimate the relative camera pose between consecutive frames in a video sequence using a view synthesis formulation. These methods work by predicting the camera poses and the depth maps, then using them to warp nearly frames to a target view using the predicted camera poses and depth maps. The learning objective is defined using the photometric loss between the predicted target view and the ground-truth target view. This view synthesis formulation implicitly makes several assumptions: 1) the scene is static; 2) there is no occlusion/disocclusion between two views; 3) there is no lighting change between two views. These assumptions often fail in applications where there exists a large displacement between the source view and the target view (see Fig.~\ref{fig:intro}).

To address these limitations, we propose a new unsupervised camera pose estimation approach using compositional re-estimation. Our proposed approach is partly inspired by the inverse compositional spatial transformer network~\cite{lin2017inverse} being developed for image alignment. The idea of our approach is that instead of estimating the relative pose between two frames in one shot, we consider the relative pose as being composed of a sequence of smaller camera poses. These smaller camera poses are estimated in a recurrent manner. The advantage of this compositional re-estimation is that we can decompose the problem of estimating the camera pose with a large displacement into several smaller ones, where each smaller problem satisfies the assumption made by the view synthesis formulation of unsupervised camera pose estimation.    

This paper makes several contributions. We propose a new compositional re-estimation approach that decomposes the camera pose estimation into a sequence of smaller pose estimation problems. Although the idea of compositional re-estimation has been used for image alignment~\cite{lin2017inverse}, this is the first work using this idea for deep visual odometry. Our model can be trained end-to-end in an unsupervised learning setting. Experimental results show that our method significantly outperforms other state-of-the-art approaches.

%% file: RelatedWork_V2.tex
\section{Related Work}
In this section, we review several lines of research closely related to our work.

\noindent{\bf Structure from Motion}: Simultaneous estimation of structure and motion is a long-standing and fundamental problem in computer vision. Traditional approaches rely on geometric constraints extracted from monocular feed to estimate motion. They commonly start with feature extraction and matching, followed by geometric verification \cite{saxena2006learning,torralba2002depth,saxena20083}. They are effective and powerful, yet computationally expensive and only focus on salient features. They also need high-quality images, and the results can drift over time due to factors such as low texture, stereo ambiguities, occlusions and complex geometry. Recently, learning-based methods have become popular and raised the bar on the performance~\cite{wang2017deepvo,kendall2015posenet,kendall2016modelling,melekhov2017image}. DeepVO~\cite{wang2017deepvo} performs end-to-end visual odometry. PoseNet~\cite{kendall2015posenet} learns 6 Degree-of-Freedom (6DOF) pose regression from monocular RGB images. Encoder-decoder style Hourglass networks have also been proposed to perform localization~\cite{melekhov2017image}.  Tang et al. \cite{tang2018ba} present BA differentiabl layer to bridge the gap between classic and deep learning methods. They minimize the feature-metric difference of aligned pixels. On the other hand, our focus is on leveraging recurrent architecture in direct method.

\noindent{\bf Depth Estimation}: Increasing availability of single view datasets \cite{geiger2013vision,Silberman,li2018megadepth} has made it possible to have significant improvement in depth prediction.  Supervised deep networks \cite{eigen2014depth,liu2018depth,liu2016learning, fu2018deep,xu2017learning,laina2016deeper,kumar2018depthnet,atapour2018real,ummenhofer2017demon} have achieved a promising performance and a variety of architectures have been proposed. Eigen et al. \cite{eigen2014depth} demonstrate the capability of deep models for single view depth estimation by directly inferring the final depth map from the input image using two scale networks. Liu et al.\cite{liu2018depth,liu2016learning} formulate depth estimation as a continuous conditional random field learning problem. Laina et al. \cite{laina2016deeper} propose the Huber loss and a newly designed up-sampling module. Kumar et al. \cite{kumar2018depthnet} demonstrate that recurrent neural networks (RNNs) can learn spatiotemporally accurate monocular depth prediction from a video. Supervised techniques are limited due to the difficulty of collecting expensive ground truth information and impractical in applications as they often require data collection process different from the target robotic deployment platform.

\noindent \textbf{Warping-based View Synthesis}: Rethinking depth estimation as an image reconstruction task allows to alleviate the need for ground-truth labels. Self-supervised approaches for structure and motion borrow ideas from warping-based view synthesis. The core idea is to supervise depth estimation by treating view-synthesis via rigid structure from motion as a proxy task. Recently, unsupervised single image camera pose estimation and depth estimation techniques have shown remarkable progress \cite{li2018undeepvo,godard2017unsupervised,zhou2017unsupervised,vijayanarasimhan2017sfm,luo2018every,flynn2016deepstereo,wang2018learning}.  These methods are mostly based on the photometric error which uses a Lambertian assumption. Garg et al. \cite{garg2016unsupervised} train a network for monocular depth estimation using a reconstruction loss over a stereo pair with  Taylor approximation to make the model fully differentiable. Godard et al.\cite{godard2017unsupervised} further improve the results by introducing symmetric left-right consistency criterion and better stereo loss functions. Zhou et al. \cite{zhou2017unsupervised}  propose a temporal reconstruction error that is computed using temporally aligned snippets of monocular images to deal with the limitation of having stereo images. The camera pose is unknown and needs to be estimated together with depth. The learning loss is obtained by combining a depth estimation network with a pose estimation network. This leads to the loss of absolute scale information in their predictions. This is solved by Li et al. \cite{li2018undeepvo} who combine both spatial and temporal reconstruction losses to directly predict the scale-aware depth and pose from stereo images. Proposed by Mahjorian et al. \cite{mahjourian2018unsupervised}, geometric constraints of the scene are enforced by an approximate ICP based loss. On the other hand, Yin et al. \cite{Yin_2018_CVPR} jointly learns monocular depth, ego-motion and optical flow from video sequences. To handle occlusion and ambiguities, an adaptive geometric consistency loss is proposed to increase robustness towards outliers and non-Lambertian regions. Geometric features are extracted over the predictions of individual modules and then combined as an image reconstruction loss. 
Last but not least, Wang et al. \cite{wang2018learning} address scale ambiguity through a compositional unit which requires Jacobian calculation. 

\noindent \textbf{Compositional and Transformer Networks}: Spatial transformer networks~\cite{jaderberg2015spatial} are developed to resolve the ambiguity of spatial variations for classification. Jaderberg et al.~\cite{jaderberg2015spatial} propose a novel strategy for integrating image warping in neural nets. Inverse compositional spatial transformers~\cite{lin2017inverse} further extends this work to remove the boundary artifacts introduced by STNs based on intuitions from the \textit{Lucal \& Kanade} algorithm~\cite{lucas1981iterative} that propagates warp parameters rather than image intensities.

%% file: Method_V6_SH.tex
\section{Our Approach}
\begin{figure}[t]
\begin{center}
   \includegraphics[width=0.70\linewidth]{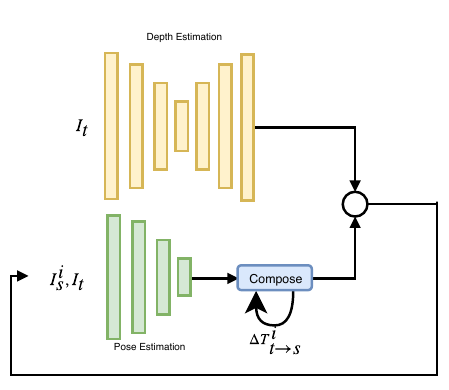}
\end{center}
\caption{The re-estimation process consists of the pose estimation network, the depth estimation network and compositional variables which keep track of the transformations. The circle indicates the inverse warping process. The recursive arrow shows the warped sources passed to the pose net for the next step.}
\label{fig:overview-recurrent}
\end{figure}

The basic components of our method are illustrated in Fig. \ref{fig:overview-recurrent}. The input to our model consists of $N$ consecutive frames in a video denoted as $<I_1,I_2,...,I_N>$. We consider one frame $I_t$ as the target frame (also known as target view or target image) and the remaining frames $I_s$ ($1\leq s\leq N, s\neq t$) as the source frames (also known as source views or source images). Our model consists of a depth network, a pose estimation network, and a warping module. The depth network produces a per-pixel depth map $D_t$ of the target frame. The pose estimation network learns to iteratively produce camera relative pose  $T^{i}_{t \to s}$ (parameterized as a 6 DoF vector representing the transformation) between the target frame $I_t$ and source frames $I_s$ where $i$ is the index of the iteration. At each iteration, we also maintain a warped source image denoted as $I^{i}_{s}$. This warped source image is obtained by applying the transformation $T^{i}_{t \to s}$ on the source image $I_s$. In other words, the pose estimation network takes a target view $I_t$ and $N$ source views $I^{i-1}_{s}$ at the $i$-th iteration as its input. It then produces $\Delta T^{i}_{t \to s}$. This transformation is combined with previous transformations $T^{i-1}_{t \to s}$ from earlier iterations to be used for warping $I_s$ (original source frames) by incorporating the depth map $D_t$ and camera intrinsics $K$ (see Sec.~\ref{sec:warping}). Let $r$ be the number of iterations of this re-estimation process. The loss function is defined in the last step of the process where $i=r$. The entire process is explained as an algorithm in the supplementary material.

\subsection{Compositional Re-estimation}
\label{re-estimation}
The goal of the compositional re-estimation module is to estimate the transformation $T^{r}_{t \to s} \in SE(3)$  from the target frame to a set of source frames. Instead of estimating the transformation in one shot, we use an iterative process that estimates this transformation incrementally. In each iteration $i$, we estimate an incremental transformation $\Delta T^{i}_{t \to s} \in SE(3)$. We use $T^{i}_{t \to s}$ to denote the transformation after the $i$-th iteration. $T^{i}_{t \to s}$ can be obtained by adding the effect of $\Delta T^{i}_{t \to s}\in SE(3)$ to the transformation matrix $T^{i-1}_{t \to s}$ from the previous iteration, i.e.
\begin{equation}
T^{i}_{t \to s} = \Delta T^{i}_{t \to s} \oplus T^{i-1}_{t \to s}
\label{eq:T_compose}
\end{equation}
where $T^{0}_{t \to s}$ includes rotation, translation. It is initialized by transformation zero and the rotation identity matrix and a row of 0 and 1 to make the matrix squared, here, $\oplus$ denotes a matrix multiplication operator. Let $r$ be the number of this compositional re-estimation steps, $T^{r}_{t \to s}$ will be used as the final transformation.

\begin{figure*}[t]
\begin{center}
   \includegraphics[width=0.90\textwidth]{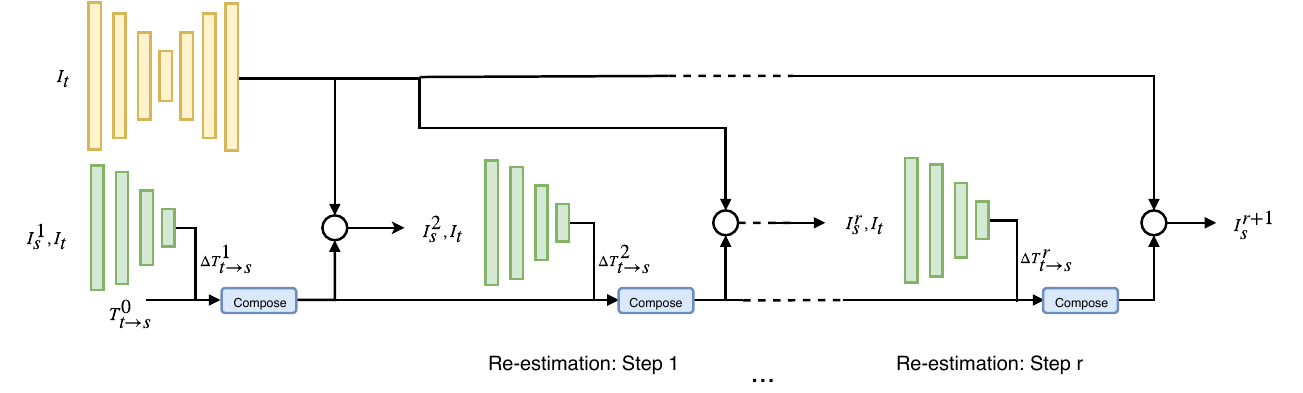}
\end{center}
   \caption{Our process is unfolded over time steps. The pose estimation network (green) estimates $\Delta T^{i}_{t \to s}$ in every steps by receiving $I^{i-1}_s$ and $I_t$. $\Delta T^{i}_{t \to s}$ is then composed to create the final $T^{r}_{t \to s}$. The loss functions will be calculated only in the last step. Warped source views $I^{r+1}_s$ from transformation $T^{r}_{t \to s}$  will be used for calculating the loss. 
}
\label{fig:overview-unfold}
\end{figure*}

The intuition behind this process is that by obtaining $T^{r}_{t \to s}$ from $\Delta T^{i}_{t \to s}$ ($i=1,2,...,r$), we allow the model to solve the camera pose estimation problem by splitting it into simpler pieces. Since each step in this process only needs to estimate a small amount of transformation, the assumptions commonly made in camera pose estimation algorithms are more likely to hold. We can unfold this process of compositional re-estimation over time steps as depicted in Fig. \ref{fig:overview-unfold}. 

\subsection{Warping Module}\label{sec:warping}   
In each estimation step $i$, a warped view $I^i_{s}$ is generated by projecting each pixel $p_t$ in the target view $I_t$ to the corresponding position $p_{s}$ in the source view (for each source view in $I^{i-1}_{s}$) and inversely warp them. This process is done for each estimation step $i \in \{1,...,r\}$. Since the process is the same throughout these time steps, we explain this warping module in one time step.

\begin{figure}[t]
\begin{center}
   \includegraphics[width=0.6\linewidth]{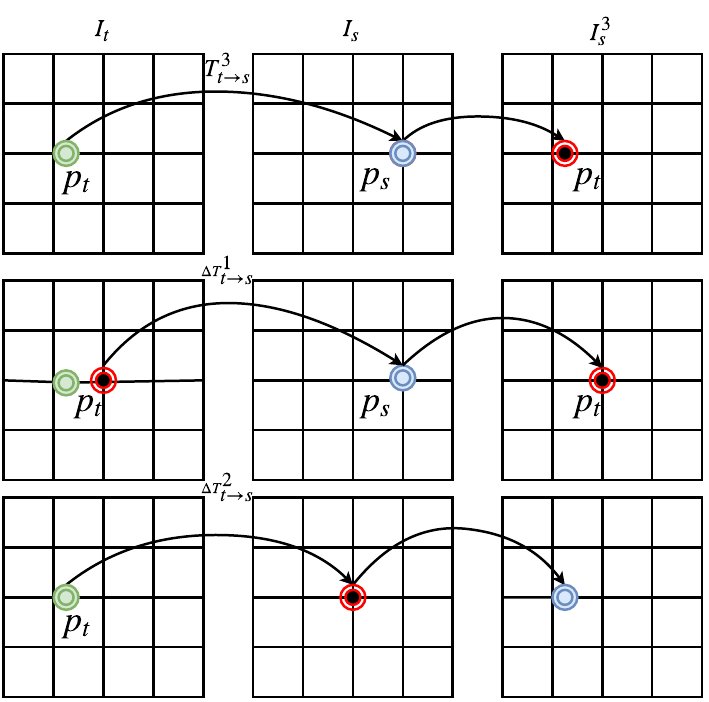}
\end{center}
\caption{The impact of two steps re-estimation is illustrated. The 2nd and 3rd rows are decompositions of the 1st row. The 1st row shows how transformation $T^{2}_{t \to s}$ leads to warping $p_s \in I_s$ to $p_t \in I_t$. It consists of 2 steps of estimation. In the first step (2nd row), the pixel $p_s$ is warped to $p_t$, but the transformation is not exactly correct. The next step (3rd row) corrects the mistake of the previous step by adding a complementary transformation to the previous step. As a result, $T^{2}_{t \to s}$ is obtained which is a true transformation from the target view to the source view. Note that although we estimate $T^{2}_{t \to s}$, we inversely warp source views to target view by the inverse of this transformation. 
}
\label{fig:projection}
\end{figure}

As shown in Fig. \ref{fig:projection}, each pixel $p_t \in I_t$ must be mapped to the corresponding $p_s \in I^{i-1}_s$. This process requires the camera intrinsics $K$, the estimated depth $D_t$ and transformation $T^{i}_{t \to s}$ (see ~Eq. \ref{eq:project}). Each $p_s \in I^{i-1}_s$ is warped to position $p_t \in I_t$ to produce $I^{i}_s$.
\begin{equation}
p_s \sim KT^{i}_{t \to s} D_{t}(p_{t})K^{-1}p_{t}
\label{eq:project}
\end{equation}
In the above equation, $K$ is a matrix of camera intrinsics and $D_{t}(p_{t})$ is the corresponding depth of $p_t$ and $T^{i}_{t \to s} \in$ SE(3).

Since some pixels are not mapped to regular grids, we reconstruct the value of $p_t$ with respect to the projection by a weighted sum of pixel neighbourhood through bilinear interpolation (Eq. \ref{eq:interpolation}) similar to \cite{zhou2017unsupervised}.

\begin{equation}
I^{i}_s(p_t) = \sum\limits_{i \in {t,b}, j \in {l,r}} w^{i,j}I^{i}_s(p_{s}^{i,j}) 
\label{eq:interpolation}
\end{equation}
In this equation, t,b,l and r denote top,bottom,left and right.

\subsection{Training Losses}
Training the re-estimation process requires a supervision signal in the form of a loss function. This loss function consists of four main components. 

\noindent{\bf Photometric Difference ($\mathcal{L}_{ph}$):} This loss function plays a vital role in our framework. Like \cite{zhou2017unsupervised,Yin_2018_CVPR,mahjourian2018unsupervised}, $\mathcal{L}_{ph}$ is an $L1$ loss between the warped source views $I^{r+1}_s$ and the target view:

\begin{equation}
\mathcal{L}_{ph} =  \sum\limits_{I \in I^{r+1}_s} \sum\limits_{p} \left | I_{t}(p) - I(p) \right |
\label{eq:photometric}
\end{equation}
where $p$ represent a pixel in an image.

\noindent{\bf Multi Scale Dissimilarity:} This term is known as DSSIM which was firstly used in \cite{Yin_2018_CVPR}. It is resilient to outliers as well as being differentiable. It calculates the dissimilarity in multi-scales of the $I^{r+1}_s$ and $I_t$. We incorporate this term with the photometric loss to form a rich dissimilarity loss. Therefore, we define it as follows:
\begin{equation}
    \mathcal{L}_{d} = \sum\limits_{i=1}^{n} \sum\limits_{I \in I^{r+1}_s} \dfrac{1 - SSIM(I,I_t)}{2}
\end{equation}
where $n$ denotes the number of scales in the prediction.

\noindent{\bf Smoothness:} This term keeps sharp details by encouraging disparities to be locally smooth. It mainly contributes to the quality of the disparity map. As most of the work on monocular depth estimation such as \cite{Yin_2018_CVPR}, we find this term very helpful in our method. We defined this term as $\mathcal{L}_{s}$.

\noindent{\bf Principled Mask:} The term of principled mask refers to an attention mechanism which ensures that out of bound pixels do not contribute to the loss function. This term is used in \cite{zhou2017unsupervised,mahjourian2018unsupervised}. In our work, this mask only contributes to the last step (r) of estimation. In order to avoid the trivial attention of zero for all pixels, we also use a regularization term ($\mathcal{L}_{reg}(E)$) in \cite{zhou2017unsupervised} in our loss function on the mask. As a result, the final photometric term in our loss function is as follows:

\begin{equation}
\mathcal{L}_{ph} = \sum\limits_{I \in I^{r+1}_s} \sum\limits_{p}  E(p) \left | I_{t}(p) - I(p) \right |
\label{eq:explainability-mask}
\end{equation}
where $E_s$ is pixel-wise predicted principled mask for the target and source and $p$ denotes a pixel.

Putting all the pieces together, the final loss function for training our model is then computed as a weighted summation of aforementioned loss functions:
\begin{equation}
\mathcal{L}_{final} = \lambda_{ph}\mathcal{L}_{ph} +\lambda_{d}\mathcal{L}_{d} + \lambda_{s}\mathcal{L}_{s} + \lambda_e \sum\limits_{i=1}^{n} \mathcal{L}_{reg}(E^i)
\label{eq:final-loss}
\end{equation}
where $\lambda_{ph}$, $\lambda_{s}$, $\lambda_{d}$ and $\lambda_e$ are loss weights.
Note that following \cite{zhou2017unsupervised}, the final loss is computed over different scales.

Since our method estimates the relative pose in multiple steps in a recurrent manner, the vanishing gradient may become an issue. To overcome this, we use residual connections and memory mechanisms in our model shown in Fig.~\ref{fig:overview-unfold}. The depth estimation network has residual connections to every differentiable warping module to alleviate the vanishing gradient problem. On the other hand, \textit{compose} $\in SE(3)$ is a variable which preserves the compositional transformation for the warping module. This variable is updated at each step so that the warping module always has access to the most updated version of transformations.

\subsection{Model Architecture}
\label{architecture-overview} 
\noindent{\bf Pose Estimation Network}: The pose estimation network is an encoder. Each layer is a convolution followed by a ReLU activation for non-linearity. The inputs to the encoder are $I_t,I^{i}_{s}$. The encoder outputs $n$ 6DOF vectors corresponding to each source view to represent camera relative poses $\Delta T^{i}_{t \to s}$ from target view $I_{t}$ to source views $I^{i}_{s}$.

In the last step of the re-estimation process, this network behaves differently, and it outputs $\Delta T^{r}_{t \to s}$ and an attention mask denoted as $E^{r}$. This attention mask is generated using a sequence of deconvolution (convTranspose) followed by sigmoid. This attention mask is used to exclude out of boundary pixels \cite{mahjourian2018unsupervised}. Note that it is acceptable that some pixels may not contribute to the loss function because they are not in target view. However, one step estimation excludes some pixels that are supposed to be in the target but are warped out of boundary due to the wrong estimation. Since we estimate the pose in multiple steps, the out of boundary pixels of ours and previous methods are different.

\noindent{\bf Depth Estimation Network}: The depth estimation network outputs the disparity map of $I_{t}$. Pixel-level depth estimation provides a rich source of information to resolve scale ambiguity of camera motion estimation \cite{zhan2018unsupervised}. In order to be consistent with both \cite{Yin_2018_CVPR} and \cite{zhan2018unsupervised}, we report the results of using both VGG-based and ResNet50-based depth estimation networks. 

%% file: Experiment_V2.tex
\section{Experiment}

We evaluate the performance of the proposed method on two complementary tasks: camera pose estimation and depth estimation. Our experiments on these tasks demonstrate that the proposed formulation leads to state-of-the-art performance for estimating the camera pose while obtaining comparable results for estimating the target frame's depth.

In the following, we first describe the implementation details of training and give details of the benchmark dataset used in the experiments. Then we present both quantitative and qualitative results. We also investigate the impact of the re-estimation process on the performance by performing ablation studies. 

\subsection{Dataset and Training Details}\label{sec:details}
\noindent\textbf{Dataset}: We evaluate our pose estimation network on the KITTI Odometry benchmark \cite{Geiger2012CVPR}. KITTI Odometry contains 22 sequences of frames recorded in street scenes from the egocentric view of the camera. Among the 22 sequences, IMU/GPS ground truth information of the first 11 sequences (seq. 00 to seq. 10) is publicly available. For the pose estimation task, we use the same training/validation splits used in \cite{zhou2017unsupervised,Yin_2018_CVPR,mahjourian2018unsupervised,zhan2018unsupervised}. For pose estimation, we train the networks on seq. 00 to seq. 08 in the official odometry benchmark of KITTI dataset. Sequence 09 and sequence 10 are reserved for evaluating the performance of camera pose estimation. Besides, we provide qualitative outputs of our approach on sequences 11 and 15, though the ground truth is not available on these sequences. For depth estimation, we use 40k frames for training and 4k for validation in order to be consistent with previous work. We evaluate the depth estimation on the split provided by Eigen et al. \cite{eigen2014depth}. It consists of 697 frames for which the depth ground truth is obtained by projecting the Velodyne laser scanned points into the image plane. 

\noindent\textbf{Training Details}: The training procedure is performed in an end-to-end fashion by jointly learning camera pose and depth estimation at the same time. Monocular frames are resized to 128 $\times$ 416 and the network is optimized by an improved variation of Adam optimizer~\cite{j.2018on}. The optimizer parameters are set to $\beta_1 = 0.9$ and $\beta_2= 0.999$. The learning rate is adjusted at $2e^{-4}$ and loss weights are set to be $\lambda_{ph} = 0.15$, $\lambda_{d} = 0.85$, $\lambda_{s} = 0.1$ and $\lambda_{e} = 0.1$. In all of our experiments, we use a batch size of 4 and set the input sequence to be 3 frames for training.

\noindent\textbf{Network Architecture}: The pose estimation network consists of 7 convolution layers followed by ReLU. The last convolution is a $1\times 1$ convolution to produce 6 DoF vectors. This 6 DoF vector corresponds to 3 Euler angles and 3-D translation which are then converted to SE(3) format for composition. In the last step of the re-estimation, the decoder of pose estimation is activated to produce the principled masks. In order to compare the depth estimation with previous work, we have experimented with using both VGG and ResNet50 as the backbone architecture in the depth estimation network. The VGG-based network is used in \cite{zhou2017unsupervised}, while the ResNet50-based network is used in \cite{Yin_2018_CVPR}.

\subsection{Monocular Pose Estimation}

As discussed before, the input to the pose estimation network is a sequence of 3 consecutive frames. We follow \cite{zhou2017unsupervised} to split the long sequences into chunks of 3 frame. The middle frame in each chunk is considered as the target frame and the other two frames as source frames. Since our work is a monocular-based system, the frames are obtained from one camera in training and testing.
In \cite{mahjourian2018unsupervised,Yin_2018_CVPR,zhou2017unsupervised}, the pose estimation network generates the camera pose vector in one step. In contrast, our approach uses the re-estimation process through composition. As a result, we achieve camera poses in a step-by-step fashion (see Sec.~\ref{re-estimation}). The performance of pose estimation is measured by the absolute trajectory error (ATE) over 3 and 5 frames snippets. Table \ref{table:vo-result-ate} compares the result of our method with other approaches.  It is noteworthy that our method does not use any external supervision signal during training. Instead, it leverages a re-estimation process which leads to a better estimation of the camera pose. Also, note that our model even outperforms other baselines that use auxiliary information. For example, ORB-SLAM \cite{murTRO2015} benefits from loop closure techniques and GeoNet \cite{Yin_2018_CVPR} utilizes the optical flow information in training. In contrast, our model does not use any of this auxiliary information. In order to evaluate the global consistency of the proposed method, we also evaluate ATE on the full trajectory which is described in \cite{sturm2012benchmark} as another measurement. Table \ref{table:vo-result-full-ate} shows the comparison with ORB-SLAM~\cite{murTRO2015} without loop closure and SFMLearner \cite{zhou2017unsupervised}.

\begin{figure}[t]
\begin{center}
   \includegraphics[width=1\linewidth]{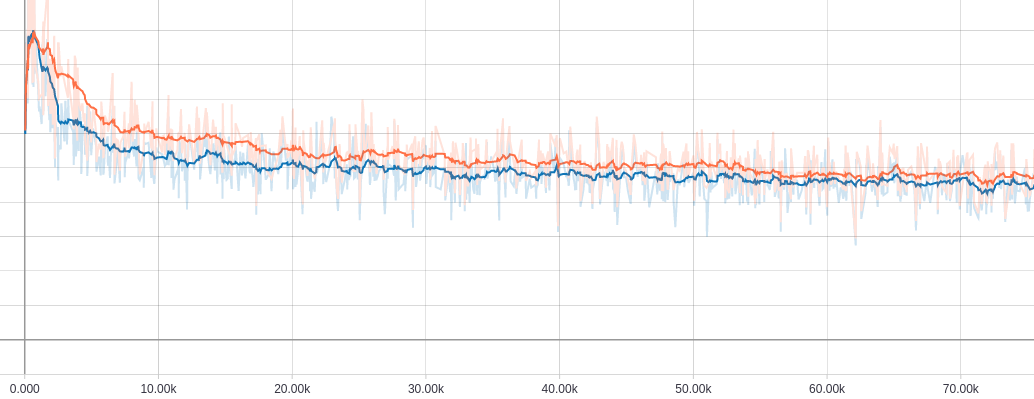}
\end{center}
   \caption{Dissimilarity loss (photometric loss + DSSIM loss) over training epochs. The loss of our approach (blue) is lower than that of the network without the re-estimation (orange) throughout the epochs. This shows that by using the re-estimation process, our model generates images that are more similar to the target frame.}
\label{fig:vo-losses}
\end{figure}

\begin{table}
\begin{center}
\begin{tabular}{|l|c|c|}
\hline
Method & seq. 9 & seq. 10 \\
\hline\hline
ORB-SLAM \cite{murTRO2015} & 0.014 $\pm$ 0.008 & 0.012 $\pm$ 0.011\\
SFMLearner \cite{zhou2017unsupervised} & 0.016 $\pm$ 0.009 & 0.013 $\pm$ 0.009\\
GeoNet \cite{Yin_2018_CVPR} & 0.012 $\pm$ 0.007 & 0.012 $\pm$ 0.009\\
3D ICP (3 frames)\cite{mahjourian2018unsupervised} & 0.013 $\pm$ 0.010 & 0.012 $\pm$ 0.011\\
EPC++(mono) \cite{luo2018every} & 0.013 $\pm$ 0.007 & 0.012 $\pm$ 0.008\\
Ours (2 steps) & \textbf{0.009} $\pm$ \textbf{0.005} & \textbf{0.009} $\pm$ \textbf{0.007}\\
\hline
\end{tabular}
\end{center}
\caption{Quantitative results for the camera pose estimation task. We compare our model with existing state-of-the-art approaches. Following prior work, we report the mean and standard deviation for Absolute Trajectory Error (ATE) over 3 and 5 snippets of sequence 9 and sequence 10 of KITTI odometry benchmark.}
\label{table:vo-result-ate}
\end{table}

\begin{table}[]
\begin{center}
\begin{tabular}{|c|c|c|c|c|}

\hline                       
Method & seq. 09 & seq. 10 \\ \hline
ORB-SLAM\cite{murTRO2015}  &    54.94         &     26.99     \\ \hline
SFMLearner \cite{zhou2017unsupervised}   &   31.21       &        28.36      \\ \hline
\textbf{Ours (2 steps)}   &  \textbf{28.38}       &       \textbf{10.25}      \\ \hline
\end{tabular}
\end{center}
\caption{Odometry evaluation on KITTI odometry benchmark sequence 09 and sequence 10. The error refers to the translational ATE error over full trajectories.}
\label{table:vo-result-full-ate}
\end{table}

\begin{figure*}[h]
	\begin{center}
		\setlength\tabcolsep{1pt}
		\def\arraystretch{0.5}
		\begin{tabular}{ccc}
			\includegraphics[height=0.60in]{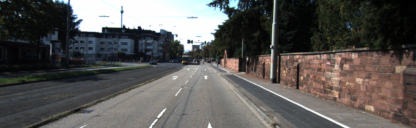}&
			\includegraphics[height=0.60in]{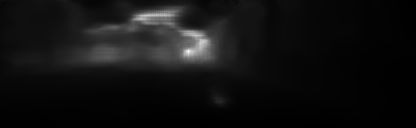}&   
			\includegraphics[height=0.60in]{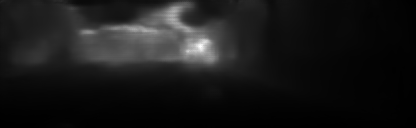}\\
			\includegraphics[height=0.60in]{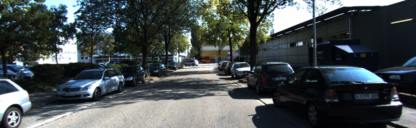}&
			\includegraphics[height=0.60in]{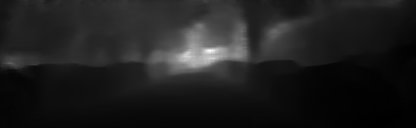}&   
			\includegraphics[height=0.60in]{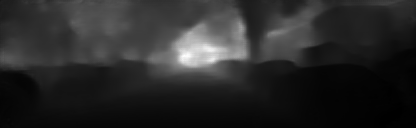}\\
			\includegraphics[height=0.60in]{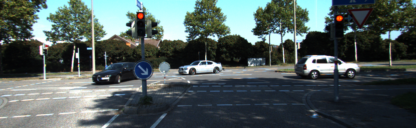}&
			\includegraphics[height=0.60in]{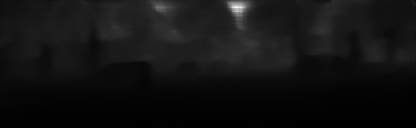}&
			\includegraphics[height=0.60in]{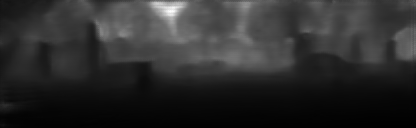}\\
			\includegraphics[height=0.60in]{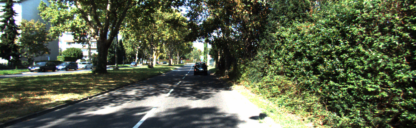}&
			\includegraphics[height=0.60in]{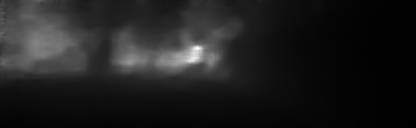}&
			\includegraphics[height=0.60in]{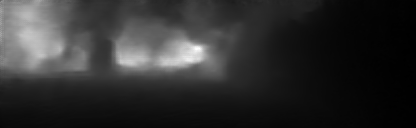}\\
		\end{tabular}
		\caption{Qualitative examples of depth estimation for one step (middle) and two steps (right) depth estimation through depth estimation network. Note that the only difference between them is the compositional re-estimation.}
		\label{fig:qualitative-depth}
	\end{center}
\end{figure*}

\subsection{Monocular Depth Estimation}\label{mono-depth}
We follow \cite{zhou2017unsupervised,Yin_2018_CVPR} in setting up the training and testing sets for the depth estimation task. More specifically, we first filter out all the testing sequence frames and frames with a very small optical flow (with magnitude less than 1) from the training set. In the end, we obtain 44540 sequences. We use 40109 of them for training and the remaining 4431 for evaluation. Note that for the task of depth estimation, the input in the training and testing phases consists of only one frame (i.e. the target frame, $I_t$).  

Similar to previous work, we multiple the predicted depth map by a scalar scale $s$ defined as $s = \textrm{median}(D_{GT})/\textrm{median}(D_{predict})$ \cite{zhou2017unsupervised}.

For a fair comparison, we compare with other monocular depth estimation approaches that use VGG and ResNet as the backbone architectures separately. Since the maximum depth in the KITTI dataset is 80 meters, we also limit the distance to 80 meters. The results are shown in Table~\ref{table:depth-result}. Although the results are comparable on the depth estimation task, our model does not outperform state-of-the-art on monocular depth estimation. This is expected since the re-estimation does not directly affect the depth estimation because it does not re-estimate the predicted depth map. This also confirms that the improvement of our method on camera pose estimation (see Table~\ref{table:vo-result-ate} and Table~\ref{table:vo-result-full-ate}) is due to the compositional re-estimation.

\begin{table*}
\begin{center}
\makebox[\textwidth]{\begin{tabular}{|c|c|c|c|c|c|c|c|c|c}
\hline
\small
Method & Supervised  & Abs Rel & Sq Rel & RMSE & RMSE log & $\delta < 1.25$ & $\delta < 1.25^{2}$ & $\delta < 1.25^{3}$\\
\hline
Cap 80m & \cellcolor{black!25}& \cellcolor{black!25}& \cellcolor{black!25}& \cellcolor{black!25}& \cellcolor{black!25}& \cellcolor{black!25}& \cellcolor{black!25}& \cellcolor{black!25}\\
\hline
Eigen et al. \cite{eigen2014depth} Coarse & Depth & 0.214 & 1.605 & 6.563 & 0.292 & 0.673 & 0.884 & 0.957\\
Eigen et al. \cite{eigen2014depth} Fine & Depth & 0.203 & 1.548 & 6.307 & 0.282 & 0.702 & 0.890 & 0.958\\
Liu et al. \cite{liu2016learning}  & Depth & 0.202 & 1.614 & 6.523 & 0.275 & 0.678 & 0.895 & 0.965\\
Godard et al. \cite{godard2017unsupervised} & Pose & 0.148 & 1.344 & 5.927 & 0.247 & 0.803 & 0.922 & 0.964\\
Zhou et al. \cite{zhou2017unsupervised} & No & 0.208 & 1.768 & 6.856 & 0.283 & 0.678 & 0.885 & 0.957\\
Zhou et al. \cite{zhou2017unsupervised} updated & No & 0.183 & 1.595 & 6.709 & 0.270 & 0.734 & 0.902 & 0.959\\
GeoNet \cite{Yin_2018_CVPR} & No & 0.164 & 1.303 & 6.090 & 0.247 & 0.765 & 0.919 & 0.968\\
ICP \cite{mahjourian2018unsupervised}  & No & 0.163 & 1.240 & 6.220 & 0.250 & 0.762 & 0.916 & 0.968\\
Ours VGG (2 steps) & No & 0.170  & 1.384 & 6.247 & 0.255 & 0.758 & 0.913 & 0.962\\
\hline
Godard et al. \cite{godard2017unsupervised} & Pose & 0.124 & 1.076 & 5.311 & 0.219 & 0.847 & 0.942 & 0.973\\
GeoNet \cite{Yin_2018_CVPR} & No & 0.153 & 1.328 & 5.737 & 0.232 & 0.802 & 0.934 & 0.972\\
Ours ResNet (2 steps) & No & 0.160  & 1.195 & 5.916 & 0.245 & 0.774 & 0.917 & 0.964 \\

\hline

\end{tabular}}
\end{center}
\caption{Quantitative results on the depth estimation task. We compare our model with other state-of-the-art monocular depth estimation approaches. Depth estimation is trained on the KITTI dataset. Evaluation is performed using the training/test split in \cite{eigen2014depth}. ``Depth'' and ``Pose'' indicate using the ground truth depth and pose as supervision during training.}
\label{table:depth-result}
\end{table*}

\subsection{Ablation Study}\label{ablation}
In order to further investigate the relative contribution of each module in our model, we perform two additional ablation studies. In the first experiment, we remove the re-estimation process in our model and train the rest of the network. We then measure the performance on the evaluation set. To do so, we set the maximum step ($r$) to 1 to assess the relative contribution of one step re-estimation process. Table \ref{table:vo-ablation1} (2nd row) shows that removing this process profoundly impacts the overall performance. The estimation accuracy drops on seq. 09 is particularly significant. This might be due to the fact that seq. 9 is more complex than seq. 10 and requires more refinement for estimating the camera pose. In the second experiment, we investigate the impact of larger displacement on the optimal number of steps. Therefore, the number of input frame is also set to be five. As it is shown in table \ref{table:vo-ablation2} and \ref{table:vo-ablation3}, the best performance on three frame snippets input is acquired by two steps estimation. However, since the displacement between source frames and target frame is larger in five frame snippets scenario, the best performance is acheived by three steps estimation.    
\begin{table}[h]
\begin{center}
\begin{tabular}{|l|c|c|}
\hline
Method & seq. 9 & seq. 10 \\
\hline\hline
ours (2 steps)& \textbf{0.009} $\pm$ \textbf{0.005} & \textbf{0.009} $\pm$ \textbf{0.007}\\
w/o re-estimation& 0.011 $\pm$  0.006 & 0.009 $\pm$  0.007\\

\hline
\end{tabular}
\end{center}
\caption{Results of ablation study of the proposed method on the pose estimation task. The 1st row shows the result of the network using the re-estimation process for 2 steps. The 2nd row shows the performance when removing it.}
\label{table:vo-ablation1}
\end{table}

\begin{figure}[h]
\begin{center}
   \includegraphics[width=0.73\linewidth]{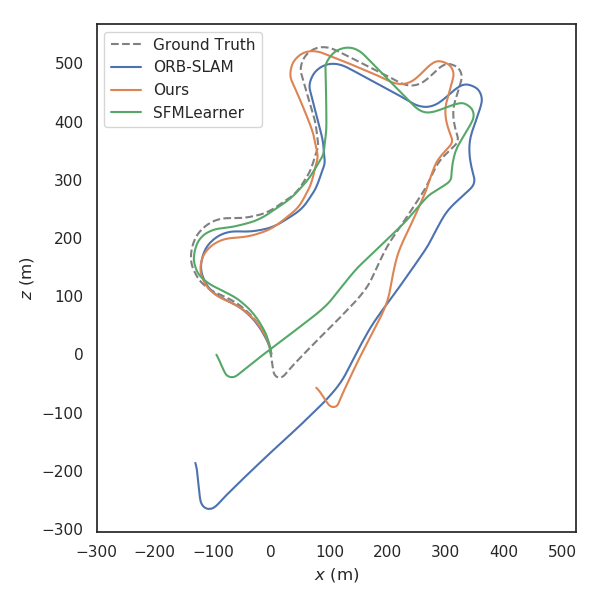}
\end{center}
\caption{Full trajectories of our method (solid orange), SFMLearner \cite{zhou2017unsupervised} (solid blue), ORB-SLAM~\cite{murTRO2015} (solid green) on the sequence 9 of KITTI Visual Odometry benchmark. Ground truth is shown in the dotted gray line.}
\label{fig:vo-results9}
\end{figure}

\begin{figure}[!h]
\begin{center}
   \includegraphics[width=0.73\linewidth]{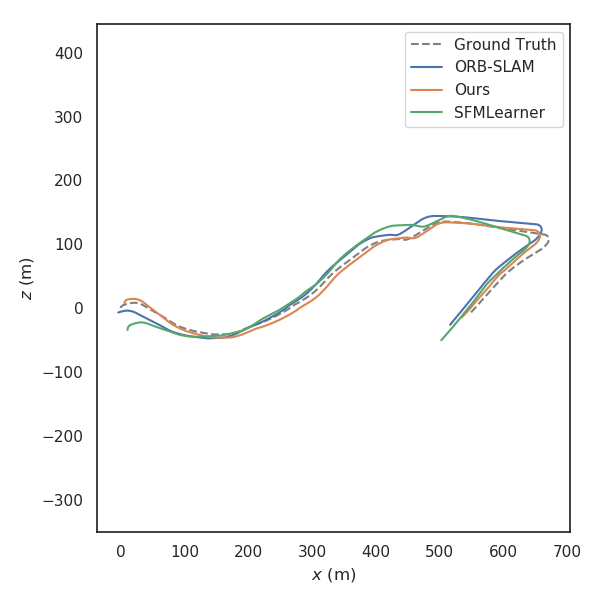}
\end{center}
   \caption{Full trajectories of our method(solid orange), SFMLearner \cite{zhou2017unsupervised} (solid blue), ORB-SLAM~\cite{murTRO2015} (solid green) on the sequence 10 of KITTI Visual Odometry benchmark. Ground truth is shown in the dotted gray line.}
\label{fig:vo-results10}
\end{figure}

Another important aspect of our method is that it leads to better image reconstruction.  In Fig. \ref{fig:vo-losses}, we visualize the re-construction loss (photometric and DSSIM) over training epochs to show how our method is better at re-construction than the baseline after a few epochs. We can see a noticeable gap between the loss of our model and the model without the re-estimation process.

\begin{table}[h]
\centering
\small
\begin{tabular}{|c|c|c|c|}
\hline
                   & \multicolumn{1}{c|}{1 step} & \multicolumn{1}{c|}{2 steps} & \multicolumn{1}{c|}{3 steps} \\ \hline
3 frames &   0.011 $\pm$ 0.006    & 0.009 $\pm$ 0.005   &     0.009 $\pm$ 0.006            \\ \hline
5 frames &  0.015 $\pm$ 0.007 &   0.014 $\pm$ 0.007      &    0.013 $\pm 0.007$                 \\ \hline
\end{tabular}
\vspace{0.1cm}
\caption{The role of the re-estimation process for 3 frame snippets and 5 frame snippets inputs on sequence 9 of KITTI odometry benchmark.}
\label{table:vo-ablation2}
\end{table}

\begin{table}[h]
\centering
\small
\begin{tabular}{|c|c|c|c|}
\hline
         &   \multicolumn{1}{c|}{1 step}     & \multicolumn{1}{c|}{2 steps} & \multicolumn{1}{c|}{3 steps} \\ \hline
3 frames &     0.009 $\pm 0.007$  &   0.009 $\pm$ 0.007    &    0.009 $\pm 0.009$          \\ \hline
5 frames &     0.014 $\pm$ 0.008     &  0.013 $\pm$ 0.008   &      0.013 $\pm$ 0.007   \\ \hline
\end{tabular}

\vspace{0.1cm}
\caption{The role of the re-estimation process for 3 frame snippets and 5 frame snippets inputs on sequence 10 of KITTI odometry benchmark.}
\label{table:vo-ablation3}
\end{table}

\subsection{Qualitative Experiment}

We provide qualitative examples for camera ego-motion estimation as the main contribution of this paper. We visualize the full trajectories on sequence 9 and 10 (Fig. \ref{fig:vo-results9} and \ref{fig:vo-results10}, respectively). Compared with \cite{zhou2017unsupervised}, our trajectories are visually better and closer to ground truth. To further demonstrate the impact of the re-estimation process, we also show the performance of our method on official test sequences (seq. 11 and seq. 15) of KITTI in supplementary material. In addition to this, we demonstrate the impact of the re-estimation process on depth estimation network in Fig. \ref{fig:qualitative-depth}. 

%% file: Conclusion.tex
\section{Conclusion and Future work}
In this paper, we have proposed a novel technique for learning to estimate camera ego motion step by step in an unsupervised deep visual odometry framework. Instead of estimating the camera pose in one pass, our method estimates the camera pose in an iterative fashion. Our method provides a new approach to address the problem of large displacement in consecutive frames. Experimental results on benchmark dataset show that our proposed method outperforms existing state-of-the-art approaches in camera pose estimation. 

%% file: aknowledegment.tex
\section{Acknowledgments}
This work was supported by a grant from NSERC. We thank NVIDIA for donating some of the GPUs used in this work.